\documentclass{article} %
\usepackage{iclr2025_conference,times}

\usepackage{amsmath,amsfonts,bm}

\def\eqref#1{equation~\ref{#1}}

\def\1{\bm{1}}

\DeclareMathAlphabet{\mathsfit}{\encodingdefault}{\sfdefault}{m}{sl}
\SetMathAlphabet{\mathsfit}{bold}{\encodingdefault}{\sfdefault}{bx}{n}

\usepackage{hyperref}
\usepackage{url}
\usepackage[utf8]{inputenc} %
\usepackage[T1]{fontenc}    %
\usepackage{hyperref}       %
\usepackage{url}    %
\usepackage{booktabs}       %
\usepackage{amsfonts}       %
\usepackage{nicefrac}       %
\usepackage{microtype}      %
\usepackage{xcolor} %

\usepackage{graphicx}
\usepackage{booktabs}
\usepackage{tabularx}
\usepackage{epsfig}
\usepackage{graphicx}
\usepackage{amsmath}
\usepackage{bm}
\usepackage{amssymb}
\usepackage{colortbl}
\usepackage{multirow}
\usepackage{color}
\usepackage{setspace}
\usepackage[normalem]{ulem}
\usepackage{tikz}
\usepackage{comment}
\usepackage{algorithm}
\usepackage{algpseudocode}
\usepackage{float}
\usepackage{caption}
\usepackage{adjustbox}
\usepackage{pifont}
\usepackage{subcaption}
\usepackage{rotating}
\usepackage{makecell}
\usepackage{wrapfig}
\newcommand{\cmark}{\ding{51}}%
\newcommand{\xmark}{\ding{55}}%

\definecolor{top1}{RGB}{245,152,153}
\definecolor{top2}{RGB}{253,205,154}
\definecolor{top3}{RGB}{248,244,140}

\usepackage[accsupp]{axessibility}  %

\definecolor{blgray}{gray}{0.97}
\definecolor{mygray}{gray}{.93}

\usepackage[capitalize]{cleveref}
\Crefname{section}{Section}{Sections}
\Crefname{table}{Table}{Tables}

\title{Scaling Concept With Text-Guided Diffusion Models}

\author{Chao Huang\textsuperscript{1}, Susan Liang\textsuperscript{1}, Yunlong Tang\textsuperscript{1}, Yapeng Tian\textsuperscript{2}, Anurag Kumar\textsuperscript{3}, Chenliang Xu\textsuperscript{1} \\
\textsuperscript{1}University of Rochester, \textsuperscript{2}The University of Texas at Dallas, \textsuperscript{3}Meta Reality Labs Research\\
}

\iclrfinalcopy %
\begin{document}

\maketitle

\begin{figure}[ht]
    \centering
    \includegraphics[width=0.99\linewidth]{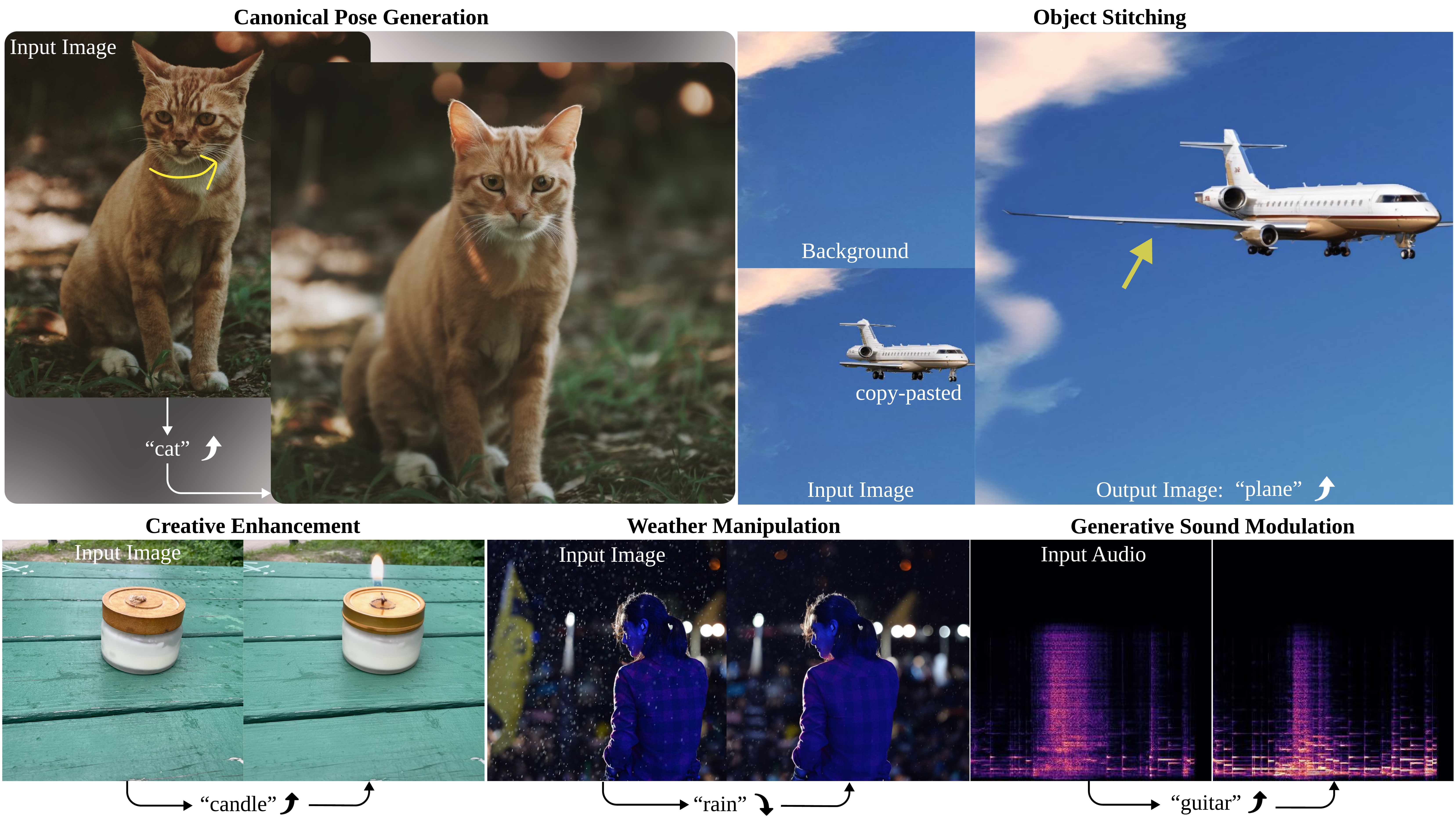}
    \caption{\textbf{Applications of ScalingConcept.} We showcase various zero-shot applications across image and audio modalities, highlighting the surprising effects of scaling concepts up or down, including non-trivial tasks such as canonical pose generation and sound modulation, among others.}
    \label{fig:teaser}
\end{figure}

\begin{abstract}

Text-guided diffusion models have revolutionized generative tasks by producing high-fidelity content from text descriptions. They have also enabled an editing paradigm where concepts can be replaced through text conditioning (e.g., \textit{a dog $\rightarrow$ a tiger}). In this work, we explore a novel approach: instead of replacing a concept, can we enhance or suppress the concept itself? Through an empirical study, we identify a trend where concepts can be decomposed in text-guided diffusion models. Leveraging this insight, we introduce \textbf{ScalingConcept}, a simple yet effective method to scale decomposed concepts up or down in real input without introducing new elements.
To systematically evaluate our approach, we present the \textit{WeakConcept-10} dataset, where concepts are imperfect and need to be enhanced. More importantly, ScalingConcept enables a variety of novel zero-shot applications across image and audio domains, including tasks such as canonical pose generation and generative sound highlighting or removal. 
Our project page is available here: \url{https://wikichao.github.io/ScalingConcept/}.

\end{abstract}

\section{Introduction}

Derived from non-equilibrium thermodynamics, diffusion models \citep{sohl2015deep} have achieved remarkable success in content generation tasks. By defining a Markov chain that progressively injects random noise into data and learning the reverse process, diffusion models generate new content iteratively from random noise. This generation paradigm has been successfully applied across various domains, including image generation~\citep{GLIDE,dalle2,imagen,stablediffusion}, video generation~\citep{ho2022video,make-a-video,tune-a-video,text2video-zero,animatediff,chen2024videocrafter2,sora}, and audio generation~\citep{diffsound,audioldm,make-an-audio,tango,audioldm2,make-an-audio2}.

Text-guided diffusion models, in particular, have garnered significant attention for their ability to control generated content using natural language prompts.
This advancement has also enabled text-guided content editing, with several works \citep{prompt2prompt,text-inversion,dreambooth,custom-diffusion,instructpix2pix,dhariwal2021diffusion,DDIM,null-text} adapting diffusion models for this purpose. For instance, DreamBooth \citep{dreambooth} fine-tunes a text-to-image diffusion model using a few images of an object paired with a text prompt $\boldsymbol{c}$ that includes the object’s class information. Null-text Inversion \citep{null-text} addresses the reconstruction errors introduced by DDIM Inversion~\citep{DDIM} in editing tasks by updating the null-text embedding. LEDITS++~\citep{brack2024ledits++} further improves the accuracy of text-guided editing and supports multiple simultaneous edits. These methods primarily focus on addressing the challenge of replacing concepts, such as using an inversion prompt  $\boldsymbol{c}=$ \textit{``a dog''} and an editing prompt $\boldsymbol{c'}=$ \textit{``a swimming dog.''}

\begin{figure}[t]
    \centering
    \includegraphics[width=0.99\textwidth]{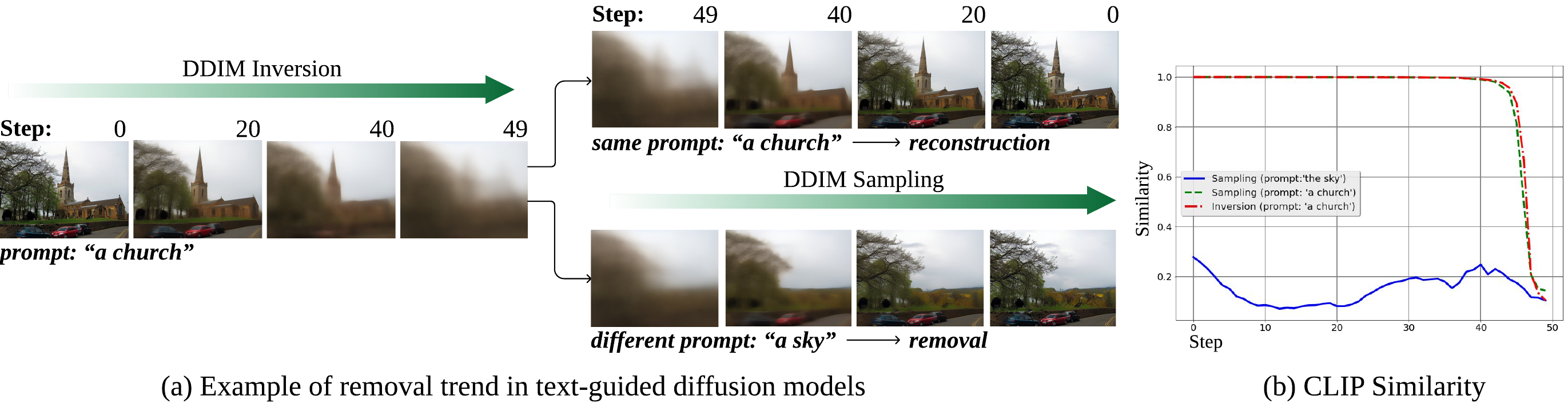}
    \caption{(a) Illustration of concept removal capability observed in the sampling process of text-guided diffusion models when conditioning on a conceptually different prompt compared to the inversion process. (b) We compute the CLIP zero-shot classification results between the classes [\textit{``a sky''}, \textit{``a church''}] and the reconstruction results at each inversion/sampling step (the total number of sampling step is 50), and report the classification accuracy of the class \textit{``a church''}. It's observed that the church object is removed from the removal branch even at the very early stages of sampling.}
    \label{fig:finding}
\end{figure}

In this work, we explore a new paradigm that moves beyond the typical editing pipeline, which generally involves replacing one concept with another. Instead, we ask: \textbf{can we scale a concept itself rather than replacing it, \textit{i.e.,} what are the effects of enhancing or suppressing a concept?} We partially answer this with a surprising observation: text-guided image diffusion models, such as Stable Diffusion~\citep{stablediffusion}, exhibit the ability to remove concepts using only text prompts. As shown in \Cref{fig:finding}, applying the prompt $\boldsymbol{c}=$ \textit{``a church''} during inversion, followed by the forward prompt $\boldsymbol{c'}=$ \textit{``a sky''}, unexpectedly removes the church and inpaints the area with content from neighboring regions. We further investigate this phenomenon by examining its \textit{scalability} and \textit{modality agnosticism}, as discussed in \cref{subsec:analysis}. Through empirical analysis, we observe that this concept removal trend exists on a scalable level and is not restricted to a single modality. It holds true across both image and audio, proving its modality-agnostic nature.

Motivated by the concept removal and reconstruction branches demonstrated in \cref{fig:finding}, we introduce our method, \textbf{ScalingConcept}, which models the difference between these two branches as a proxy for representing the concept itself. Specifically, given a concept $\boldsymbol{c}$ to be scaled, we apply an inversion technique using text-guided diffusion models to obtain the concept-sensitive latent variable $\boldsymbol{x}_T$. During the sampling process, we model the difference between the noise predictions for the prompt $\boldsymbol{c}$ (reconstruction) and the null prompt $\emptyset$ (removal). A scaling factor is incorporated to control this modeling process across different diffusion time steps.
Additionally, we introduce a noise regularization term to better balance fidelity with concept scaling. Experiments on our \textit{WeakConcept-10} dataset demonstrate that our method outperforms baseline editing-oriented approaches in concept scaling, with detailed analysis of the impact of each component.

Our zero-shot ScalingConcept method unlocks a variety of downstream applications (as shown in \Cref{fig:teaser}) without additional cost. Scaling up a concept standardizes its representation, while scaling down tends to remove it. In the image domain, this enables tasks such as canonical pose generation, object stitching, weather manipulation, and more. Our concept scaling adjusts non-standard object poses, completes stitched objects, and integrates them seamlessly with the background. It also allows for weather modifications, such as deraining or dehazing.
In the audio domain, we achieve sound highlighting by amplifying text-indicated sounds while suppressing others, and generative sound removal by decomposing audio mixtures into individual components. 

Unlike most existing diffusion-based editing methods, ScalingConcept requires no customized layers or additional training. It only relies on a text-guided inversion-forward process, making it easily reproducible with any text-guided diffusion model. Furthermore, if a more advanced text-guided diffusion model becomes available, ScalingConcept can be seamlessly integrated to achieve a wide range of applications with minimal cost.

In all, our contributions can be summarized as follows:
\begin{itemize}
    \item We conduct a comprehensive empirical study of the concept removal phenomenon across both image and audio domains, laying the groundwork for moving beyond the traditional concept replacement approach. 
    \item We propose the ScalingConcept method, which models the difference between concept reconstruction and removal, incorporating a scaling factor and noise regularization to provide precise control over concept scaling during the diffusion process.
    \item We validate the effectiveness of ScalingConcept quantitatively on the newly collected \textit{WeakConcept-10} dataset and demonstrate its versatility through a variety of zero-shot applications in both image and audio domains, such as canonical pose generation, object stitching, weather manipulation, sound highlighting, and generative sound removal—all achieved without the need for additional fine-tuning. 
\end{itemize}

\section{Related Works}
\subsection{Text-guided Diffusion Models}

Text-guided diffusion models have set a new standard for realistic content generation across multiple domains, including images~\citep{GLIDE,dalle2,imagen,stablediffusion}, videos~\citep{ho2022video,make-a-video,tune-a-video,text2video-zero,animatediff,tang2024cardiff,sora}, and audio~\citep{diffsound,audioldm,make-an-audio,tango,audioldm2,make-an-audio2}. A key factor behind their success is the deep integration of language understanding into the content generation process. For instance, the GLIDE model~\citep{GLIDE} introduced text-conditional diffusion models that enable controlled image synthesis, while DALL-E 2~\citep{dalle2} employed a two-stage approach leveraging joint CLIP embeddings~\citep{clip} to capture semantic information from text inputs. Similarly, Imagen~\citep{imagen} showcased the efficacy of large pre-trained language models like T5~\citep{t5} in encoding text prompts for image generation tasks. Latent Diffusion Models, such as Stable Diffusion~\citep{stablediffusion}, further optimized the diffusion process by performing it in the latent space, enhancing both efficiency and generation quality.
The success observed in the image domain has extended to other modalities. For instance, methods like the Video Diffusion Model (VDM)\citep{ho2022video}, Make-A-Video\citep{make-a-video}, AnimateDiff~\citep{animatediff}, and VideoCrafter~\citep{chen2023videocrafter1} have adapted image diffusion models to generate videos from text. In the audio domain, methods such as AudioLDM~\citep{audioldm}, Make-An-Audio~\citep{make-an-audio}, and TANGO~\citep{tango} have achieved similar breakthroughs, demonstrating the versatility of diffusion models across modalities. The success of these models is rooted in their ability to learn robust text-to-modality associations, showing that textual concepts can be effectively translated into various types of content. In our work, we build upon these associations, introducing a novel approach to leverage text-guided diffusion models across multiple modalities for the purpose of concept scaling.

\subsection{Text-guided Editing with Diffusion Models}
Text-guided content editing using diffusion models has rapidly advanced in recent years. Methods such as DreamBooth~\citep{dreambooth}, Null-text Inversion~\citep{null-text}, and InstructPix2Pix~\citep{instructpix2pix} have introduced techniques to fine-tune and control diffusion models for specific editing tasks. These approaches primarily focus on replacing or modifying objects within an image by manipulating inversion techniques and applying further learning. For instance, DreamBooth~\citep{dreambooth} allows for text-guided personalization of diffusion models by fine-tuning them with a small number of images, and OAVE~\citep{liang2024language} extends it to audio-visual editing. Null-text Inversion~\citep{null-text} addresses reconstruction errors in concept editing through optimizing null-text embeddings. A more recent approach, LEDITS++~\citep{brack2024ledits++}, introduces an efficient inversion method to produce high-fidelity results with fewer diffusion steps while supporting multiple simultaneous edits. 
In contrast to these methods, which focus on concept personalization or replacement, our approach introduces a novel paradigm: concept scaling. We explore how diffusion models can systematically remove or amplify concepts across different modalities, unlocking a wider range of applications.

\section{Method}

In this section, we first review the foundational knowledge of text-guided diffusion models and diffusion inversion techniques in \Cref{subsec:background}, which form the basis of our method. Next, we present an empirical analysis of the trend of concept removal observed in text-guided diffusion models in \Cref{subsec:analysis}. Finally, in \Cref{subsec:decomposition_diffusion}, we introduce our novel approach, ScalingConcept, which allows flexible control over the strength of the target concept in real input data.

\subsection{Preliminary}
\label{subsec:background}

\textbf{Text-guided diffusion models.} Text-guided diffusion models have gained significant attention for their success in generating realistic images, audio, and video from text prompts. Their key strength lies in accurately capturing text-to-X associations, where X refers to any modality. Taking an image as an example, the process typically begins using an autoencoder such as VQ-GAN~\citep{esser2021taming} to project an input into a latent vector $\boldsymbol{x_0}$.
During diffusion, Gaussian noise is progressively added to the latent feature, resulting in a random noise vector $\boldsymbol{x_T}$. In the denoising phase, a noise prediction network $\epsilon_\theta$ learns to estimate the noise added at each step. Text-guided diffusion models use a text condition $\boldsymbol{c}$, usually derived from text embeddings like CLIP~\citep{clip}, to guide the sequential denoising process. The learning objective is defined as: 
\begin{equation}
    \label{eq:diffusion}
    \ell_{simple}  = ||\epsilon - \epsilon_\theta(\boldsymbol{x_t}, \boldsymbol{c}, t)||, 
\end{equation} 
where $\epsilon$ is the Gaussian noise added at timestep $t$.

\begin{figure}[t]
    \centering
    \includegraphics[width=0.8\textwidth]{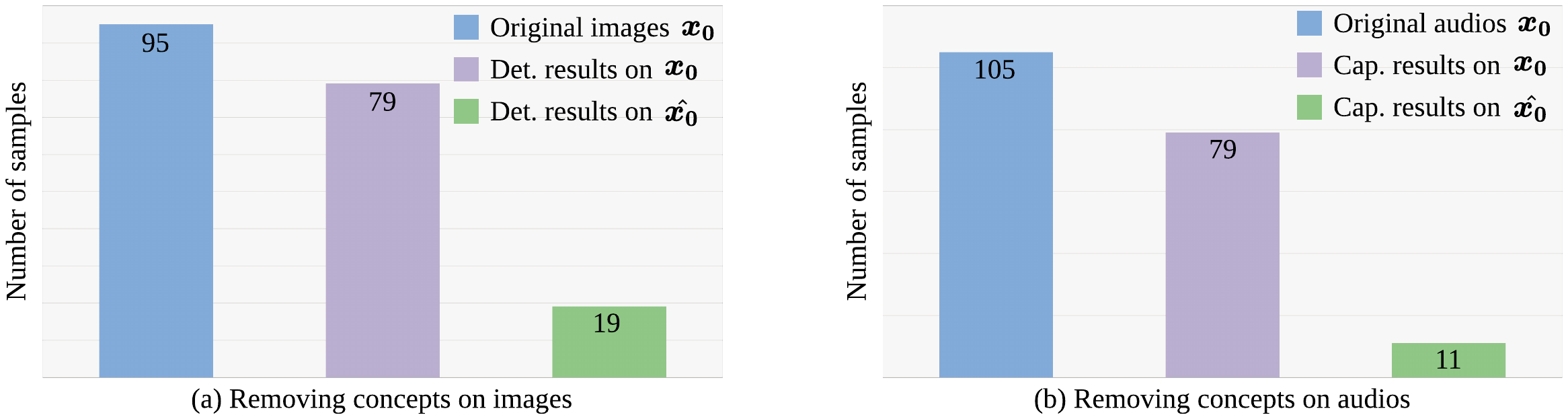}
    \caption{Analysis of the trend of concept removal. We erase target concepts from given images and audio clips using the proposed inversion and sampling process. We report the number of samples with target concepts before and after concept removal.}
    \label{fig:statics}
\end{figure}

\textbf{Inversion technique.} Inversion techniques are commonly used in generative diffusion models to enable the editing of real content~\citep{xia2022gan,text-inversion,null-text}. Typical inversion methods, such as DDIM inversion~\citep{dhariwal2021diffusion,DDIM}, convert an input latent $\boldsymbol{x_0}$ into a noisy latent variable $\boldsymbol{x_T}$, which can then be used to reconstruct $\boldsymbol{x_0}$ or perform edits. Specifically, DDIM inversion leverages its deterministic sampling process:
\begin{equation}
    \label{eq:ddim}
   \boldsymbol{x_{t-1}} = \sqrt{\frac{\bar{\alpha_{t-1}}}{\bar{\alpha_t}}} \boldsymbol{x_t} + \left(  \sqrt{\frac{1}{\bar{\alpha_{t-1}}}-1} - \sqrt{\frac{1}{\bar{\alpha_t}} - 1} \right) \epsilon_\theta (\boldsymbol{x_t}, \boldsymbol{c}, t),
\end{equation}
with $\{\bar{\alpha_t}\}_{t=0}^T$ as a predefined noise schedule. This process iteratively denoises $\boldsymbol{x_T}$ to recover $\boldsymbol{x_0}$. Due to ODE formulation, it can be reversed, with small steps, to obtain the inversion (denoted as $f^{inv}( \boldsymbol{x_{t}}, \boldsymbol{c}, t)$):
\begin{equation}
	\label{eq:inversion}
    \boldsymbol{x_{t+1}} = \sqrt{\frac{\bar{\alpha_{t+1}}}{\bar{\alpha_t}}} \boldsymbol{x_t} + \left(  \sqrt{\frac{1}{\bar{\alpha_{t+1}}}-1} - \sqrt{\frac{1}{\bar{\alpha_t}} - 1} \right)  \epsilon_\theta (\boldsymbol{x_t}, \boldsymbol{c}, t),
\end{equation}
 thereby estimating the noisy latent $\boldsymbol{x_T}$ from $\boldsymbol{x_0}$. Starting with this $\boldsymbol{x_T}$, the sampling process can be guided by arbitrary text conditions. However, DDIM inversion is limited by cumulative errors at each step, which deviate the path toward the correct latent noise. Several methods, such as DDPM inversion~\citep{huberman2024edit} and ReNoise~\citep{garibi2024renoise}, have been proposed to improve the inversion process.

\subsection{Empirical Analysis on the Concept Removal}
\label{subsec:analysis}

 \Cref{eq:inversion} and \Cref{eq:ddim} define a pair of destruction and reconstruction processes that have been successfully applied in prior research for concept editing. Given an input $\boldsymbol{x_0}$, the inversion process extracts its latent variable counterpart $\boldsymbol{x_T}$, and the reverse process generates an edited output where the original concept $\boldsymbol{c}$ is modified to $\tilde{\boldsymbol{c}}$, allowing for various types of editing.
A case study of this paradigm is shown in \Cref{fig:finding}, where we perform an inversion with the prompt \textit{``a church,''} that branches into two sampling paths: (1) using the same prompt, ``a church,'' to reconstruct the image as expected, and (2) using the prompt \textit{``a sky.''} Interestingly, in the second path, the church is removed, and the vacated area is inpainted with content related to the surrounding context, even from the first sampling step.
We hypothesize that this removal effect is due to the interplay between cross- and self-attention mechanisms in diffusion models. During inversion, the noise estimator $\epsilon_\theta$ relies heavily on cross-attention to incorporate context from $\boldsymbol{c}$, leading to strong modifications in regions associated with the concept $\boldsymbol{c}$. However, during sampling, when the prompt \textit{``a sky''} provides no useful context for reconstructing the church, self-attention becomes dominant, leading to the church's removal.

\textbf{Does the concept removal trend appear at scale?} To determine if the above concept removal phenomenon is isolated or consistent across a broader dataset, we replicate the process using 95 samples from 10 common classes in the COCO dataset~\citep{lin2014microsoft}. For each image $\boldsymbol{x_0}$, we apply DDIM inversion with the prompt ``[class].'' After obtaining the noisy latent variable $\boldsymbol{x_T}$, we use a null prompt $\emptyset$ during sampling to convert $\boldsymbol{x_T}$ back into an image $\hat{\boldsymbol{x}}_0$. This process mirrors that in \Cref{fig:finding}, aiming to remove the concept ``[class]'' from the input image. To evaluate whether the concept is successfully removed, we use Grounding DINO \citep{liu2023grounding} to detect the presence of the ``[class]'' object in both $\boldsymbol{x_0}$ and $\hat{\boldsymbol{x}}_0$. The results, presented in \Cref{fig:statics}, show that the target concept ``[class]'' is successfully removed in 80\% of the images. This confirms that the concept removal capability exists at scale, rather than being limited to an individual sample.

\textbf{Does the concept removal apply to other modalities?}
To investigate this, we conduct a similar experiment with audio, another common modality. Using the AVE dataset~\citep{tian2018audio}, an audio event classification dataset containing clips from 28 sound classes, we randomly sample 5 audio clips from each class. We employ AudioLDM 2~\citep{audioldm2} to replicate the process used in the image-based experiment. To determine whether the concept is removed from the original audio clip, we use EnCLAP~\citep{kim2024enclap}, an audio captioning framework, to generate captions for both $\boldsymbol{x_0}$ and $\hat{\boldsymbol{x}}_0$. We then check whether the word ``[class]'' appears in the caption. As shown in \Cref{fig:statics}, the same trend of concept removal is observed in audio, despite its fundamentally different nature compared to images.

\textbf{Discussions.} From the empirical analysis above, we observe that starting from the same latent variable $\boldsymbol{x_{T}}$ obtained through inversion, both a reconstruction branch and a removal branch can be defined. This implicitly suggests that text-guided diffusion models have the ability to \textbf{decompose a concept}. Based on these findings, an important research question arises: can we control the divergence between these two branches to achieve concept scaling?

\subsection{Our Method: ScalingConcept}
\label{subsec:decomposition_diffusion}
Motivated by the difference between the removal and reconstruction branches, we propose \textbf{ScalingConcept}, a method designed to decompose the concept from real input and scale it up or down, effectively enhancing or suppressing the corresponding representation. Our method consists of two steps:

\noindent \textbf{Step 1: generating the scaling startpoint $\boldsymbol{x_{T}}$.}  Given a real input $\boldsymbol{x_0}$ and a concept $\boldsymbol{c}$ to scale,  represented by a text prompt such as \textit{``fire hydrant,''} we use a pre-trained text-guided diffusion model $\epsilon_\theta$ to perform sequential inversion functions as described in  \Cref{eq:inversion}:
\begin{equation}
	\boldsymbol{x_{T}} = f^{inv}( \boldsymbol{x_0}, \boldsymbol{c}, 0) \circ ... \circ f^{inv}( \boldsymbol{x_{T-1}}, \boldsymbol{c}, T-1).
\end{equation}
In our experiment, we use ReNoise~\cite{garibi2024renoise} as the inversion technique.

 \noindent \textbf{Step 2: concept scaling.}
 Starting from $\boldsymbol{x_{T}}$, we define two prompts: the first is the text prompt $\boldsymbol{c}$ used during inversion, corresponding to the reconstruction branch, and the second is the null-text prompt $\emptyset$, representing the removal branch.  The noise predictions from the two branches are denoted as $\epsilon^{\emptyset}_t = \epsilon_\theta(\boldsymbol{x_t}, \emptyset, t)$ and $\epsilon^{r}_t = \epsilon_\theta(\boldsymbol{x_t}, \boldsymbol{c}, t)$, where the superscript $r$ stands for reconstruction. We model the difference between these two branches by manipulating the difference in their noise predictions.

\begin{equation}
\Hat{\epsilon}_t = \epsilon^{\emptyset}_t + \omega_t \cdot (\epsilon^{r}_t - \epsilon^{\emptyset}_t).
\label{eq:conceptscaling}
\end{equation}

\begin{figure}[t]
	\centering
	\includegraphics[width=0.8\textwidth]{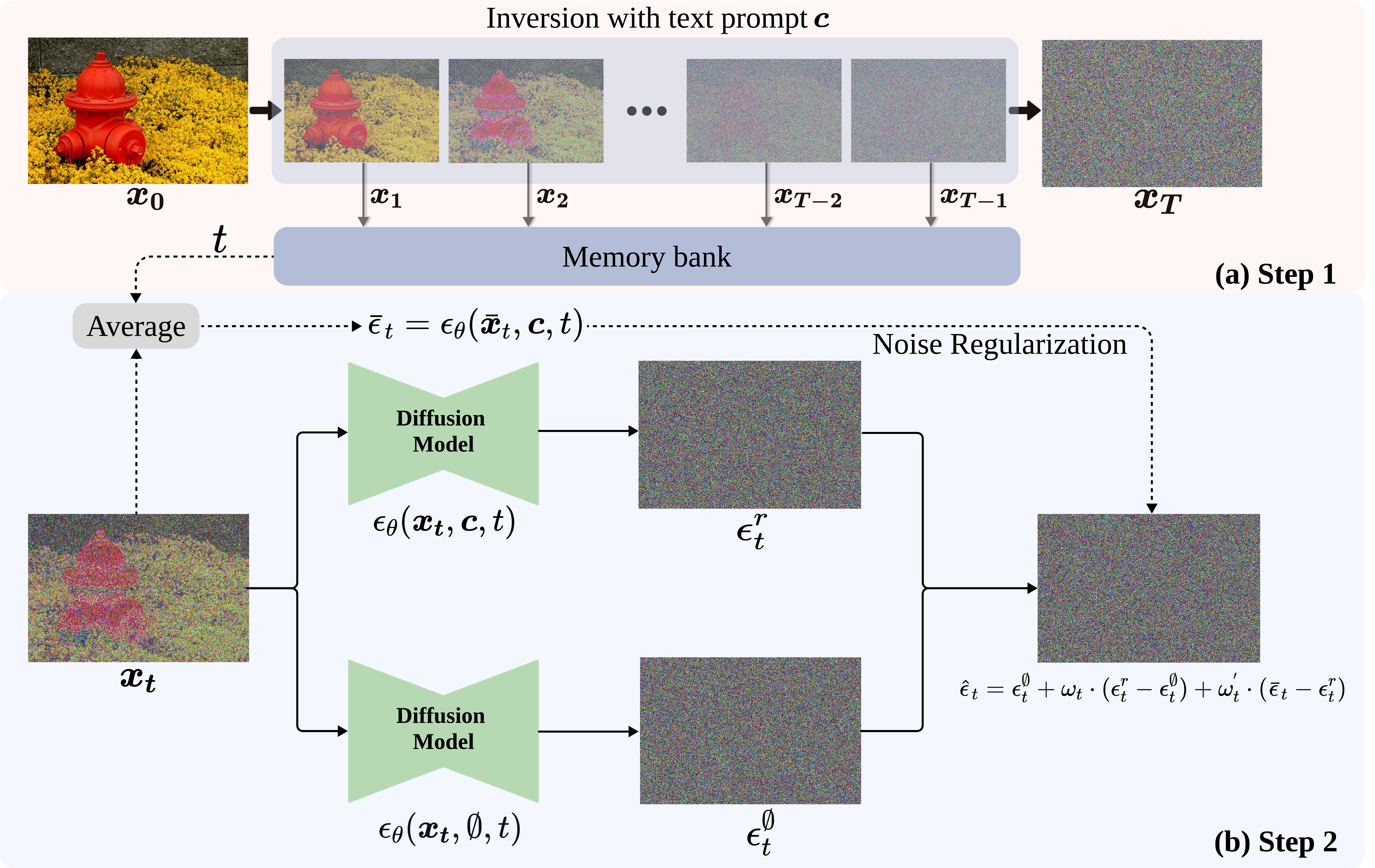}
	\caption{Overview of the ScalingConcept framework. Our method consists of two steps: 1) extracting the latent variable from $\boldsymbol{x}_0$, and 2) constructing different sampling branches and modeling the difference between them. }
        \vspace{-3mm}
	\label{fig:method}
\end{figure}

Here, we introduce a scaling factor $\omega_t$ to control the magnitude of the difference at each step $t$. Note that when $\omega_t=1$, \Cref{eq:conceptscaling} reduces to the vanilla reconstruction branch. A value of $\omega_t < 1$ suppresses the concept, while $\omega_t > 1$ enhances it. Intuitively, during the early steps of inference, the model captures coarse-grained details such as global structure and shape, whereas in the final steps, it focuses on refining high-frequency details \citep{si2024freeu}. To explore the impact of different designs for $\omega_t$, we express it as $\omega_t = \omega_{base} * \beta(t)$, where $\omega_{base}$ controls the overall strength of scaling, and $\beta(t)$ is a scheduling function within the range 0 to 1. We propose a dynamic schedule, $\beta(t) = \left(\frac{t}{T}\right)^\gamma$, where $\gamma$ controls the sharpness of scaling. This approach supports three common types of schedule: 1) Constant ($\gamma=0$),  treating the difference equally across all steps, similar to classifier-free guidance in diffusion models. 2) Linear ($\gamma=1$), reflecting a linear change in the concept's impact over time. 3) Non-linear ($\gamma \neq 0$ or $1$), allowing for dynamic adjustments of the concept's influence, depending on the value of $\gamma$.

\noindent \textbf{Noise regularization.} When $\omega_t$ is set to a very large value, the noise prediction $\Hat{\epsilon}_t$ in \Cref{eq:conceptscaling} can deviate significantly from the original input, leading to dissimilar content despite the concept being scaled --- an undesired effect. Our goal is to scale the concept while preserving the context of the original input. To address this, we introduce a noise regularization term. At each timestep $t$, we retrieve the corresponding noisy latent generated during the inversion process from the memory bank. We combine this with the current noisy latent, adjust the noise predictions using an averaging operation, and then reintroduce them into \Cref{eq:conceptscaling_reg} using the same scaling factor.
Additionally, since the forward noisy latents deviate further from the inversion latents in the later steps, we apply an early exit method to stop noise regularization when necessary. The regularized noise prediction is defined as:

\begin{equation} 
\Hat{\epsilon}_t = \epsilon^{\emptyset}_t + \omega_t \cdot (\epsilon^{r}_t - \epsilon^{\emptyset}_t) + \omega_t^{'} \cdot (\bar{\epsilon}_t - \epsilon^{r}_t),%
\label{eq:conceptscaling_reg} 
\end{equation}

\begin{equation}
\omega_t^{'} := \begin{cases}
  0 &\quad\text{if } t < t_{exit}, \\
  \omega_t &\quad\text{otherwise.} \\ 
\end{cases}
\end{equation}

In our experiment, $t_{exit}$ is empirically set to 35, out of a total of 50 sampling steps.

\begin{figure}[t]
    \centering
    \begin{minipage}[h]{0.53\textwidth}
        \centering
        \captionof{table}{Comparison of different methods for concept enhancement. Our method, ScalingConcept, achieves the best performance in terms of image quality (lower FID score), maintaining original content (lower LPIPS), and comparable concept enhancement (similar CLIP score) to other approaches.}
        \label{tab:method_compare}
        \footnotesize
        \begin{tabularx}{1\textwidth}{lXccc}
            \toprule
            Method & & FID $\downarrow$ & CLIP (\%) $\uparrow$ & LPIPS $\downarrow$ \\
            \midrule
           \rowcolor{lightgray} \textit{Input} & &313.4 & 26.9 & -  \\
            Instruct P2P & & 312.0 & 27.8 & 0.312 \\
            LEDITS++ & &274.4 & \textbf{28.6} & 0.321 \\
            Ours & &\textbf{272.2} & \textbf{28.6} & \textbf{0.291}  \\
            \bottomrule
        \end{tabularx}
    \end{minipage}
    \hfill
    \begin{minipage}[h]{0.45\textwidth}
        \centering
        \includegraphics[width=\linewidth]{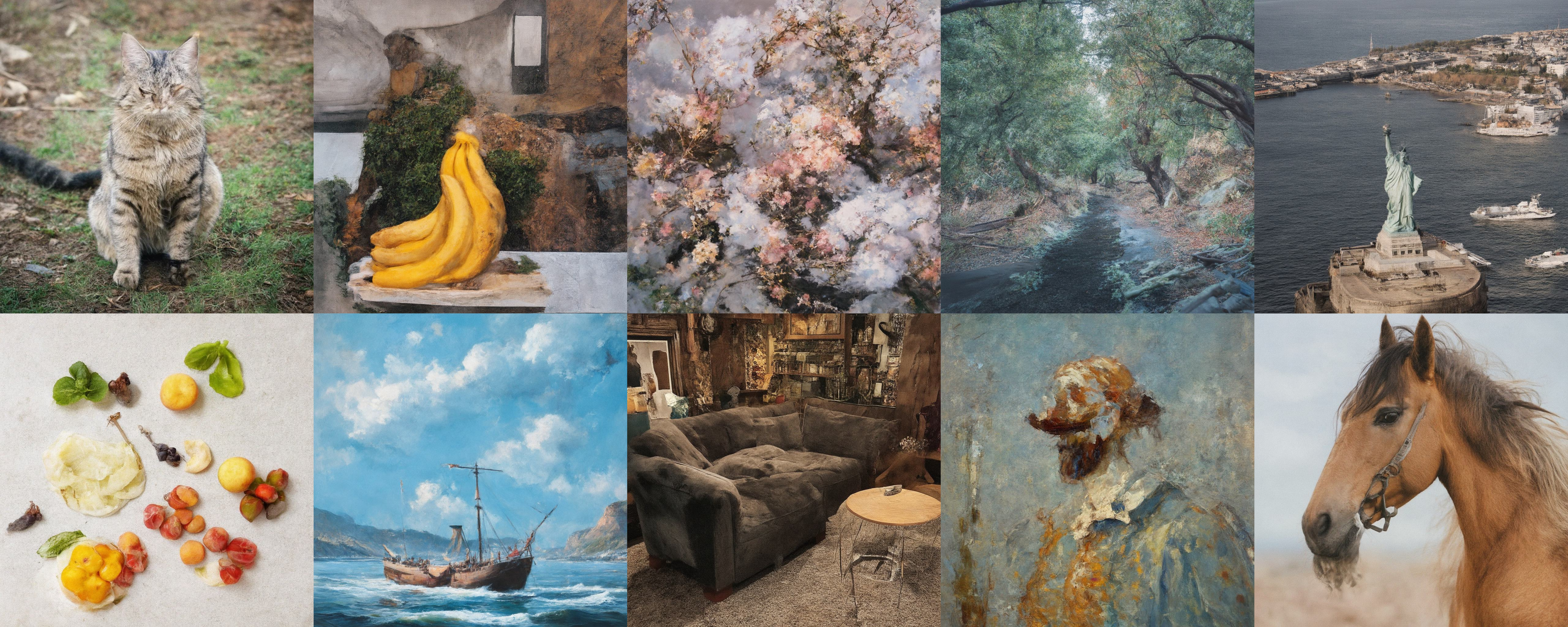}
        \caption{Overview of the WeakConcept-10 dataset. The images exhibit weak and incomplete representations of the target concepts, making them ideal candidates for evaluating concept scaling methods.}
        \label{fig:dataset overview}
    \end{minipage}
\end{figure}

\section{Experiment}
\subsection{WeakConcept-10 Dataset} To effectively test concept scaling, a dataset that supports the measurement of concept strength is crucial. However, evaluating whether a concept has been enhanced or suppressed in real inputs poses a significant challenge. To address this, we leverage Stable-Diffusion-3 (SD3) \citep{sd3}, a recently released and powerful text-guided image diffusion model, to generate images exhibiting weak concepts.
We begin by selecting 10 diverse categories, including \textit{sofa, banana, cat, flower, Van Gogh, ship, Statue of Liberty, fruits, forest,} and \textit{horse}. For each category, we generate 10 images using the prompt ``[class\_name]'' with classifier-free guidance of 1, ensuring that the generated images reflect weak representations of the target concept.
As illustrated in \Cref{fig:dataset overview}, the generated images display indistinct structures and missing details of the specified concept, making them ideal candidates for improvement through concept scaling.

\noindent \textbf{Evaluation metric.} We utilize three metrics to evaluate performance: CLIP score \citep{clip}, FID \citep{fid}, and LPIPS \citep{lpips}. The CLIP score measures the similarity between the image and the text prompt, assessing whether the target concept has been successfully enhanced. FID evaluates the overall image quality after concept scaling, while LPIPS measures the perceptual similarity between the enhanced output and the original weak input.
\begin{figure}[h]
    \centering
    \includegraphics[width=0.98\linewidth]{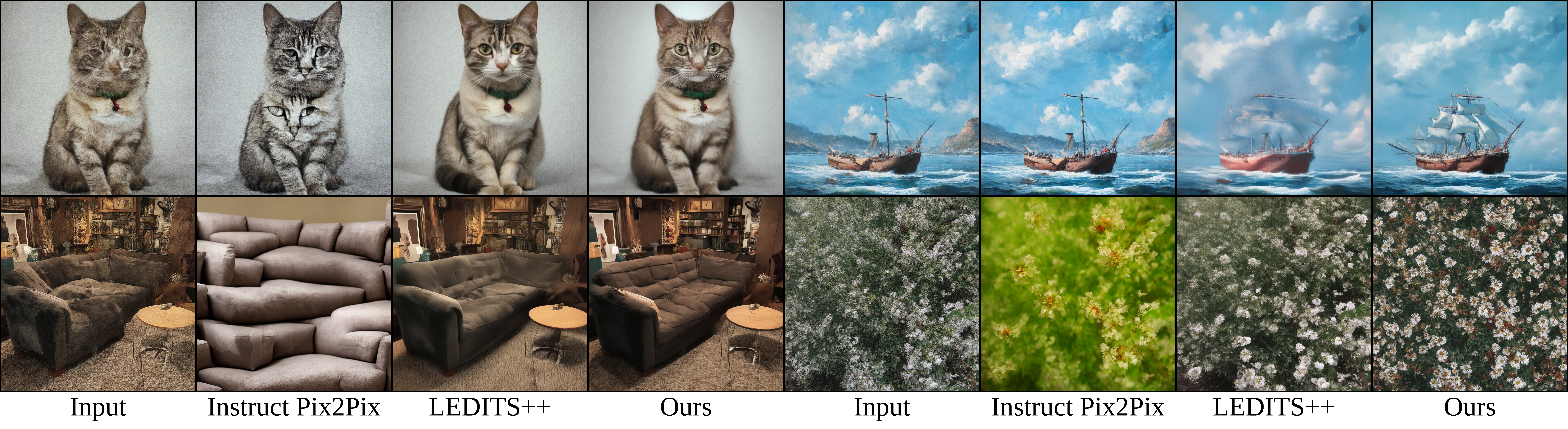}
    \caption{Qualitative comparison with baseline methods. We present input images with weak concepts from our dataset, alongside the enhanced results from two baseline approaches and our ScalingConcept method.}
    \label{fig:dataset comparison}
\end{figure}

\subsection{Main Comparison} To evaluate the effectiveness of our ScalingConcept method, we compare it against Instruct Pix2Pix \citep{instructpix2pix}, which enhances the concept using the prompt ``enhance the [concept]''. Additionally, we adapt the editing method LEDITS++ \citep{brack2024ledits++} for our experiment. While LEDITS++ is proposed to add or remove new concepts, in our case, we use it to add the existing concept ``[concept]'', effectively simulating concept enhancement.
The comparison results are presented in \Cref{tab:method_compare}. Both LEDITS++ and our method achieve comparable concept strength, as indicated by similar CLIP scores. However, our method produces superior image quality, reflected by a lower FID score, while also preserving the original context of the input. This demonstrates the effectiveness of ScalingConcept in both enhancing the concept and maintaining image fidelity. For qualitative comparison, see \Cref{fig:dataset comparison}, where our method clearly enhances the weak concept while preserving fine details in the image.

\subsection{Ablation Studies}
In \Cref{tab:ablation}, we analyze the trade-off between fidelity and generation quality by varying the value of $\gamma$ and introducing noise regularization. We set $\omega_{base} = 5$ for all the ablations. The CLIP score for all variants remains similar (28.5 - 28.7), indicating that $\omega_{base}$ effectively controls the strength of concept scaling.  Our goal is to strike a better balance between concept scaling and content preservation.

\noindent \textbf{Effect of different $\gamma$.} As $\gamma$ increases, the FID score rises, suggesting a shift from pure generation toward a balance between preserving the original content and enhancing the concept, as reflected by the corresponding improvement in the LPIPS score. In this work, we aim to scale the concept while maintaining this balance. Thus, we select a relatively large value for $\gamma$, such as 3.

\noindent \textbf{Effect of noise regularization and early exit.} Introducing the noise regularization term significantly improves the LPIPS score from 0.324 to 0.260, indicating better preservation of the original content. However, this also constrains concept scaling. When early exit is applied, both FID and CLIP scores improve, though content preservation is slightly compromised, leading to a better overall balance.

\begin{table}[!t]
    \caption{Ablation studies. We set $\omega_{base} = 5$ for all experiments and test the performance with various values of $\gamma$. Additionally, we examine the impact of noise regularization and early exit on the results.}
    \centering
    \footnotesize
    \begin{tabularx}{0.95\textwidth}{lccXccc}
        \toprule
        Configuration & Noise Regularization & Early Exit & & FID  & CLIP (\%) & LPIPS \\
        \midrule
        $\gamma=0$ (Constant) & \xmark & \xmark & & 232.9 & 28.6 & 0.397\\
        $\gamma=0.5$ (Non-linear) & \xmark & \xmark & & 238.6 & 28.7 & 0.380\\
        $\gamma=1$ (Linear) & \xmark & \xmark & & 242.0 & 28.7 & 0.368\\
        $\gamma=3$ (Non-linear) & \xmark & \xmark & & 258.1 & 28.5 & 0.324\\
        \midrule
        \multirow{2}{*}{$\gamma=3$} & \cmark & \xmark & & 282.6 & 28.5 & 0.260 \\
        & \cmark &\cmark & & 272.2 & 28.6 & 0.291\\
        \bottomrule
    \end{tabularx}
    \vspace{-4mm}
    \label{tab:ablation}
\end{table}

\section{Application Zoo}
In this section, we present the application zoo, showcasing several applications enabled by our ScalingConcept method. All results are achieved in a zero-shot manner, emphasizing the versatility and value of our approach. These applications are non-trivial and span both image and audio domains. For image tasks, we use SDXL~\cite{sdxl} as our base model, while for audio tasks, we employ AudioLDM 2~\citep{audioldm2}.

\begin{figure}[h]
    \centering
    \includegraphics[width=0.98\linewidth]{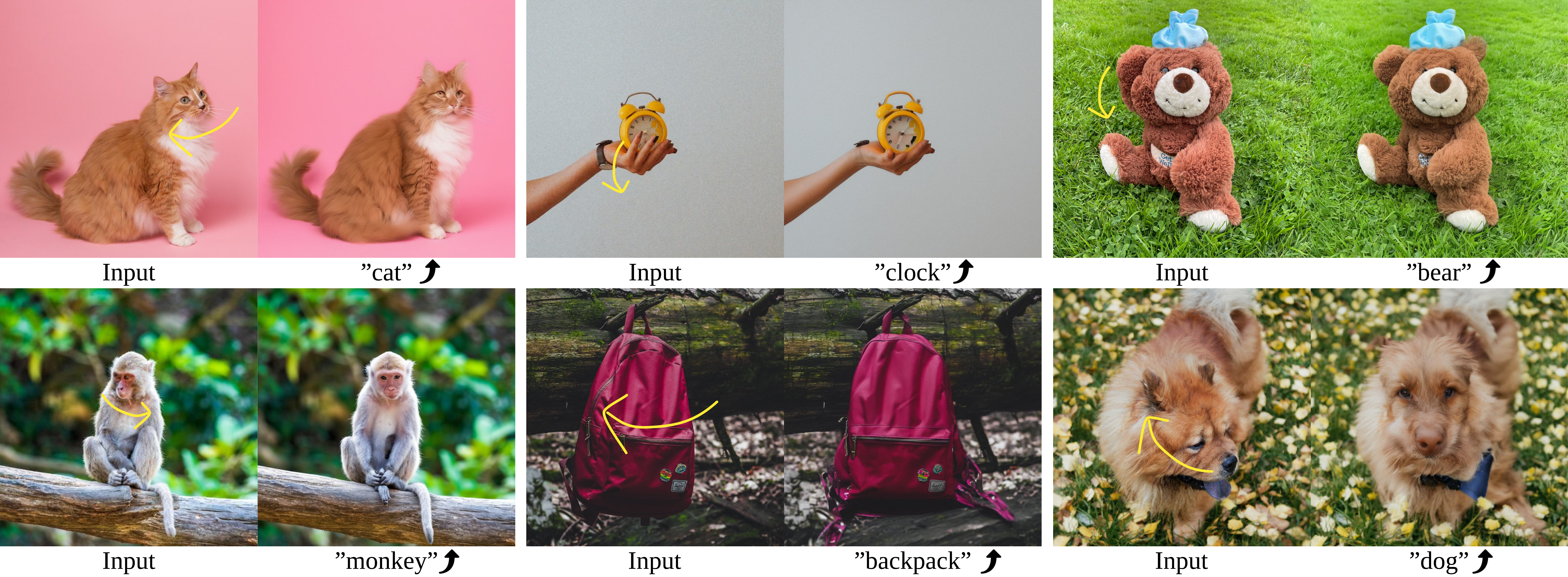}
    \caption{Canonical pose generation. By scaling up the concept of an object, our model adjusts its pose to be more complete and visible.}
    \label{fig:pose generation}
    \vspace{-6mm}
\end{figure}

\noindent \textbf{Canonical pose generation.} Our ScalingConcept method enables an interesting and non-trivial task: adjusting the pose of the subject in an image by scaling up the concept. In \Cref{fig:pose generation}, we demonstrate the effect of canonical pose generation. In the original input images, the concepts to be scaled up --- such as the cat, clock, and backpack --- are depicted in various poses. After applying concept scaling, the cat and backpack are adjusted to face forward, and the clock’s occlusion by a hand is mitigated, resulting in a more complete expression of the concept.
Across all results, scaling up the concept facilitates seamless and faithful pose adjustments, a task that is challenging even in the 3D domain but is effectively handled by our method. From a high-level perspective, scaling up the concept enhances its completeness and visibility, often resulting in front-facing orientations. This technique has potential applications in 3D tasks such as novel-view synthesis.

\begin{figure}[h]
    \centering
    \includegraphics[width=0.98\linewidth]{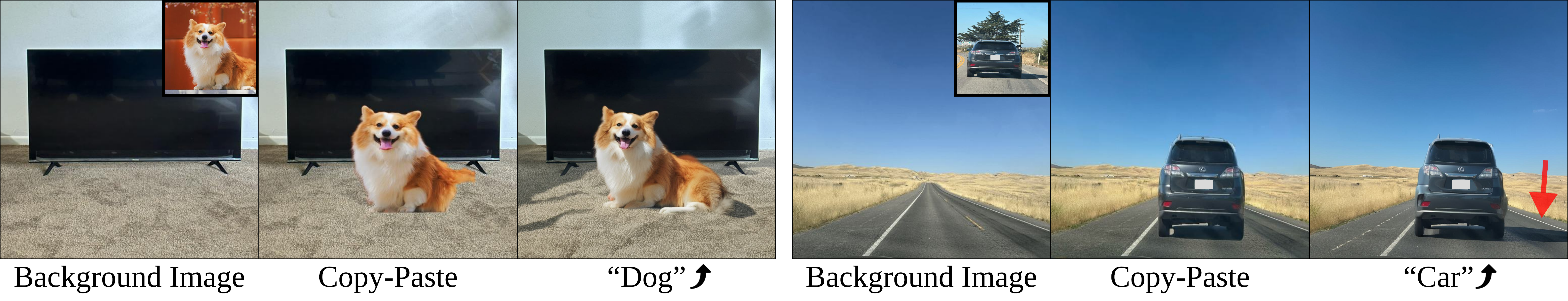}
    \caption{Object stitching. By enhancing an object's concept, our method seamlessly stitches the object and the background together, completing and harmonizing the whole image.}
    \vspace{-3mm}
    \label{fig:stitching}
\end{figure}

\noindent \textbf{Object stitching.} Another straightforward application is diffusion-based object stitching~\cite{song2022objectstitch, song2023objectstitch}. When an object is copied and pasted into a background image, scaling up the concept in the pasted image makes the object more complete. For example, in \Cref{fig:stitching}, we observe the dog being completed, lighting adjustments made, and the shadow of the car added, seamlessly blending the object with the background.

\begin{figure}[!h]
    \centering
    \includegraphics[width=0.7\linewidth]{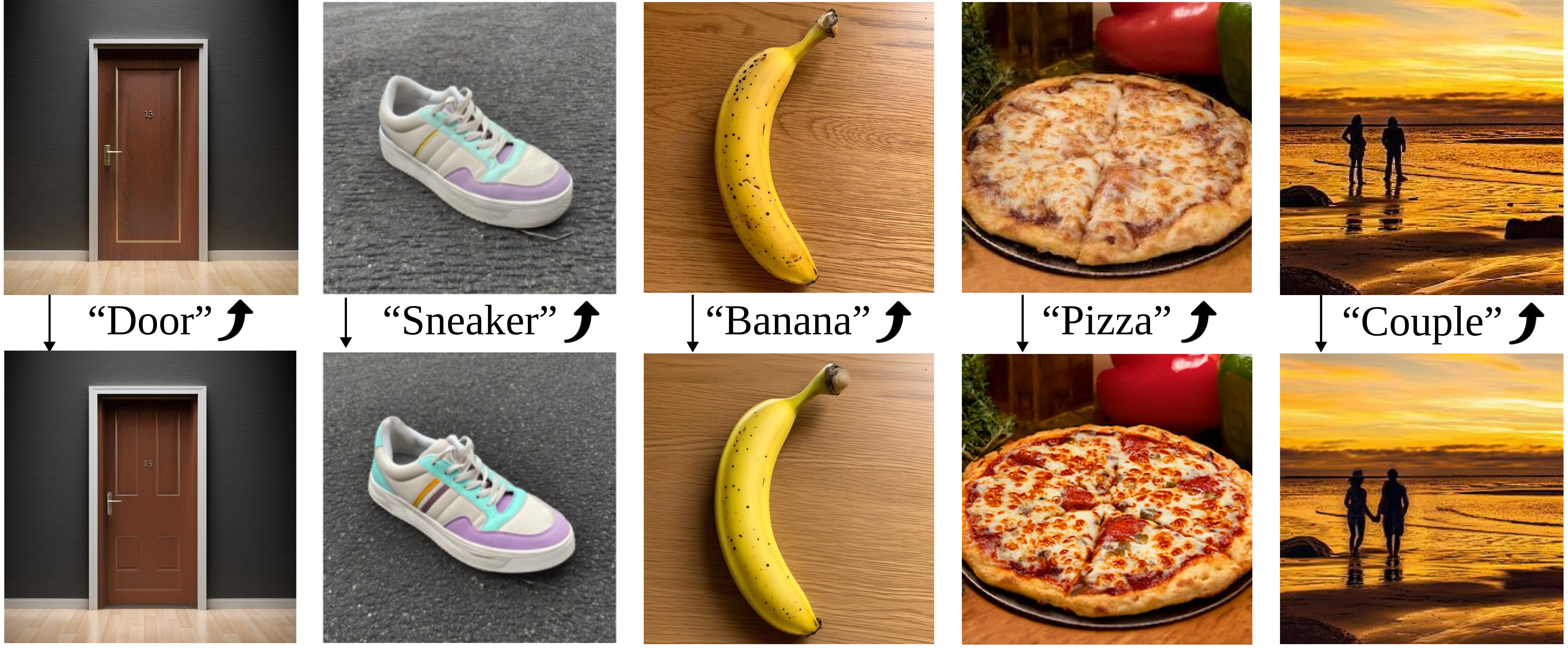}
    \caption{Creative enhancement. ScalingConcept produces ``growing'', enhancing and expanding the concept of input images.}
    \vspace{-3mm}
    \label{fig:creative enhancement}
\end{figure}
\noindent \textbf{Creative enhancement.} A more open-ended application, as shown in \Cref{fig:creative enhancement}, is creative enhancement. In this case, the effect of scaling up the concept is dependent on the specific content of the image, often producing surprising ``growing'' effects. For example, when scaling up the concept, the \textit{``couple''} transitions from standing separately to holding hands; and the \textit{``pizza''} gains additional toppings. This application is particularly useful for users who want to explore different effects by enhancing concepts in arbitrary images.

\begin{figure}[t]
    \centering
    \includegraphics[width=0.99\linewidth]{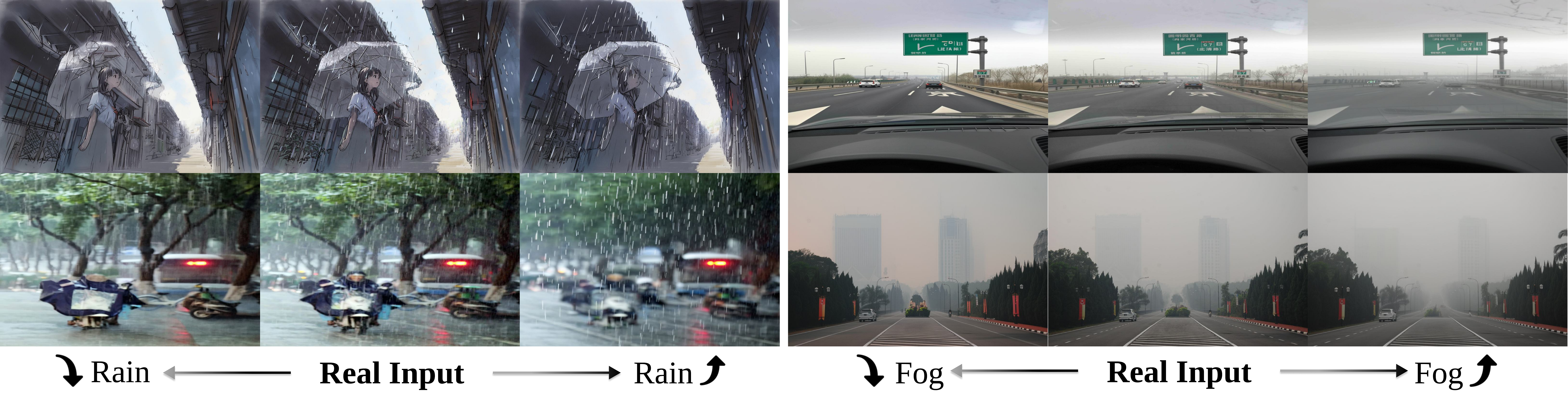}
    \caption{Weather manipulation. Our method enables both weather suppression, akin to deraining and dehazing tasks, as well as weather enhancement.}
    \label{fig:weather manipulation}
\end{figure}

\noindent \textbf{Weather manipulation.} Since our method supports both scaling up and down concepts, a practical application is weather manipulation (as shown in \Cref{fig:weather manipulation}). Scaling down can address classic weather mitigation tasks, such as deraining or dehazing, while scaling up is useful in scenarios like movie production, where specific weather conditions are needed. For example, in the movie ``The Mist,'' there would be no need to wait for naturally heavy fog --- our method can effectively enhance the fog to achieve the desired effect.

\begin{figure}[h]
    \centering
    \includegraphics[width=0.99\linewidth]{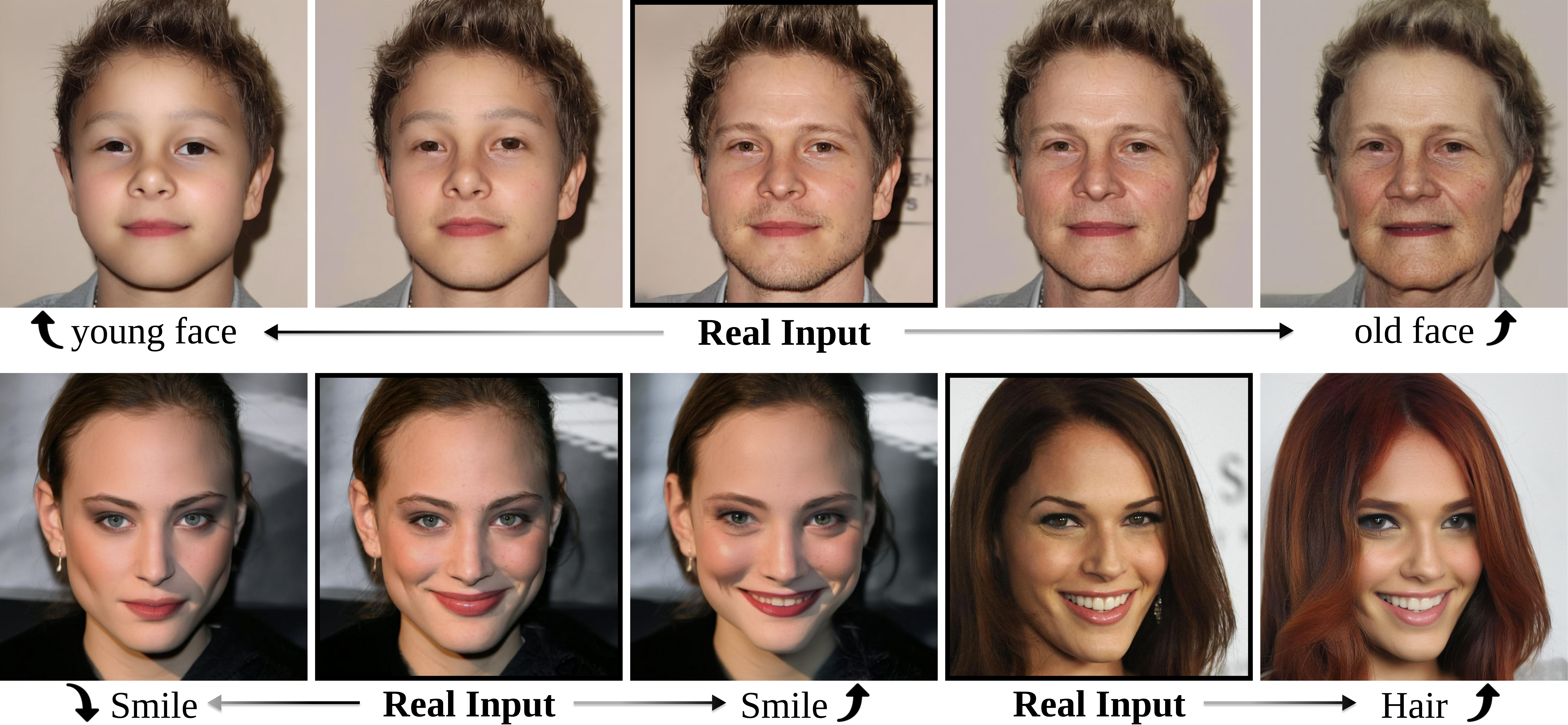}
    \caption{We present a random batch of 3 samples from CelebA-HQ~\cite{CelebA-HQ}, without cherry-picking, to demonstrate our method’s versatility in scaling different face attribute concepts.}
    \label{fig:face}
\end{figure}

\noindent \textbf{Face attribute scaling.} We extend our method to face images. In \Cref{fig:face}, we showcase popular face attribute editing tasks on examples from the CelebA-HQ~\cite{CelebA-HQ} dataset, such as adjusting age, smile, and hair. Each of these edits can be achieved by scaling the corresponding concepts, demonstrating the versatility of our method.

\begin{figure}[h]
    \centering
    \includegraphics[width=0.99\linewidth]{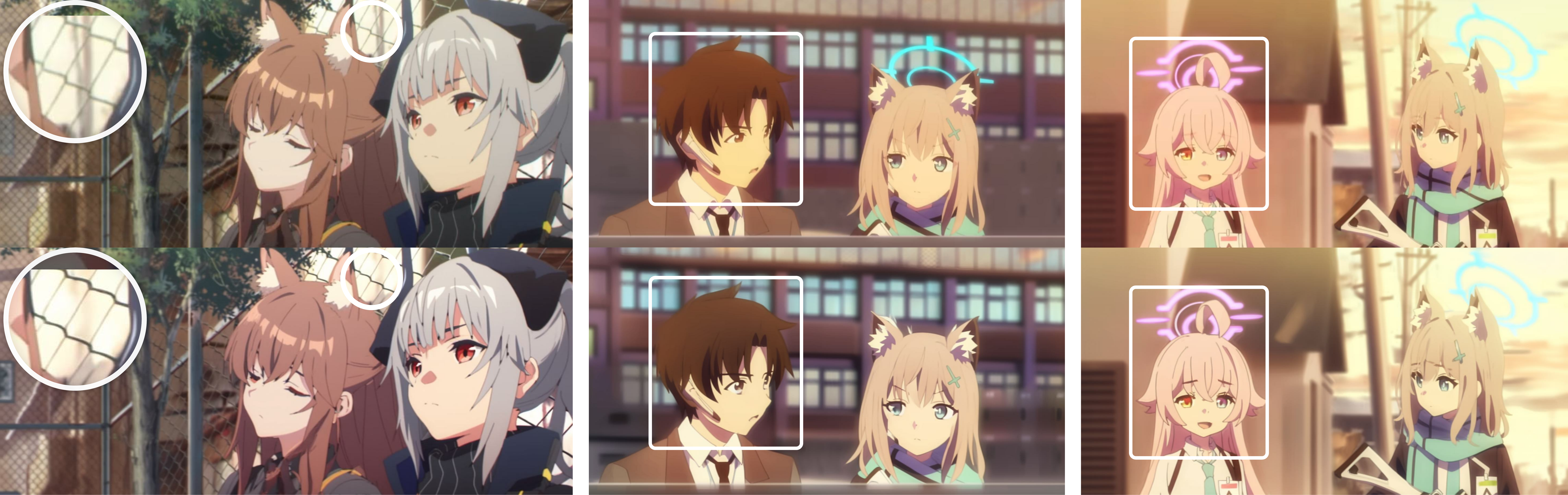}
    \caption{The top row shows screenshots from the anime \textit{``Arknights''} (Left) and \textit{``Blue archive''} (Middle \& Right).The bottom row displays the images after scaling up the ``anime'' concept, which mitigates the fuzziness and blurriness issues commonly encountered in the anime production process. }
    \label{fig:anime}
\end{figure}

\noindent \textbf{Anime skectch enhancement.} During the photography and post-production stages of anime making, cumulative errors in line processing often result in blurred lines, making the image appear fuzzy. Filters for scenes like sunsets exacerbate this issue, which cannot be resolved simply by increasing the resolution or bitrate of the anime. Using our ScalingConcept method, we process images with such issues by applying "anime" as the concept to scale up. This enhances the sketches in the image as shown in \cref{fig:anime}, leading to an overall improvement in visual clarity.

\begin{figure*}[!h]
\centering

    \includegraphics[width=\linewidth]{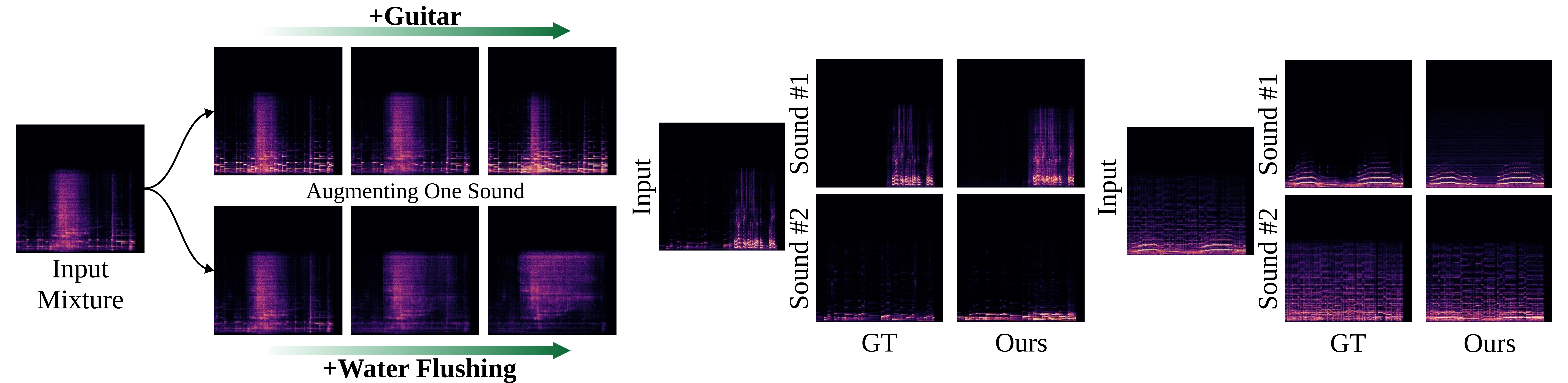}
    \centering
\caption{(Left) Sound highlighting. Our method increases the volume of a target sound and keeps the other sounds intact. (Middle \& Right) Qualitative comparison on sound separation. Our method enables sound removal through a generative model.}
\label{fig:combined_figure}
\end{figure*}

\noindent \textbf{Generative sound highlighting.} For audio applications, we introduce a new task --- sound highlighting, which involves increasing the volume of a target sound by scaling the concept using our method. As shown in \Cref{fig:combined_figure}, starting from a mixture of sounds, we can highlight either the guitar sound or the water-flushing sound, while preserving the presence of the other sounds on the track.

\noindent \textbf{Generative sound removal.} Another audio application is sound removal from an audio track, similar to sound separation~\cite{huang2023davis}, but achieved through a generative model. In \Cref{fig:combined_figure}, we use a mixture of sounds as input and scale down the concept by specifying the class of the non-target sound as the inversion prompt.

\section{Conclusion}
We propose ScalingConcept, a zero-shot concept scaling method that focuses on enhancing or suppressing existing concepts in real input data. Our method allows for user-friendly adjustments by freely tuning the scaling strength $\omega_{base}$ and the scaling schedule $\gamma$, enabling a wide range of effects. More importantly, ScalingConcept unlocks numerous non-trivial applications across various modalities, such as canonical pose generation and sound removal or highlighting. Our approach has the potential to become a valuable tool within the family of diffusion models.

This new approach to concept manipulation also comes with new challenges, particularly in defining concepts textually, setting hyperparameters, and managing potential fine-tuning needs. Current editing methods required years of refinement to address similar issues, and we anticipate that future work will successfully tackle these challenges for ScalingConcept as well.

\newpage
\bibliography{iclr2025_conference}
\bibliographystyle{iclr2025_conference}

\appendix
\section{Appendix}

\subsection{Limitations}
Despite our method presenting a zero-shot approach to scaling concepts in real inputs and achieving promising results, there are several limitations to the current method.

\noindent \textbf{Choice of hyperparameters.} In our current method, we split the scaling factor $\omega_t$ into two controlling factors: $\omega_{base}$ and the schedule $\beta(t) = \left(\frac{t}{T}\right)^\gamma$. Users can adjust $\omega_{base}$ and $\gamma$ to control the scaling strength. Although we demonstrate the effects of different components in \cref{tab:ablation} and \cref{tab:method_compare}, the optimal combination varies depending on the task, making user input non-trivial. To address this, a potential future direction is to design an automatic scaling factor that adapts to the target concept's strength, thus eliminating the need for extensive hyperparameter tuning.

\noindent \textbf{Dependence on text-to-X association.} While our method enables concept scaling with text-guided diffusion models for any modality (X), its effectiveness relies heavily on the text-to-X association. If the text prompt is not sensitive to the diffusion model -- meaning the information about the concept is not captured effectively -- the method may fail. To address this issue, incorporating concept-specific fine-tuning may be beneficial for certain edge cases.

\subsection{Is Canonical Pose Generation Easy to Achieve?}
\begin{figure*}[!h]
\centering

    \includegraphics[width=\linewidth]{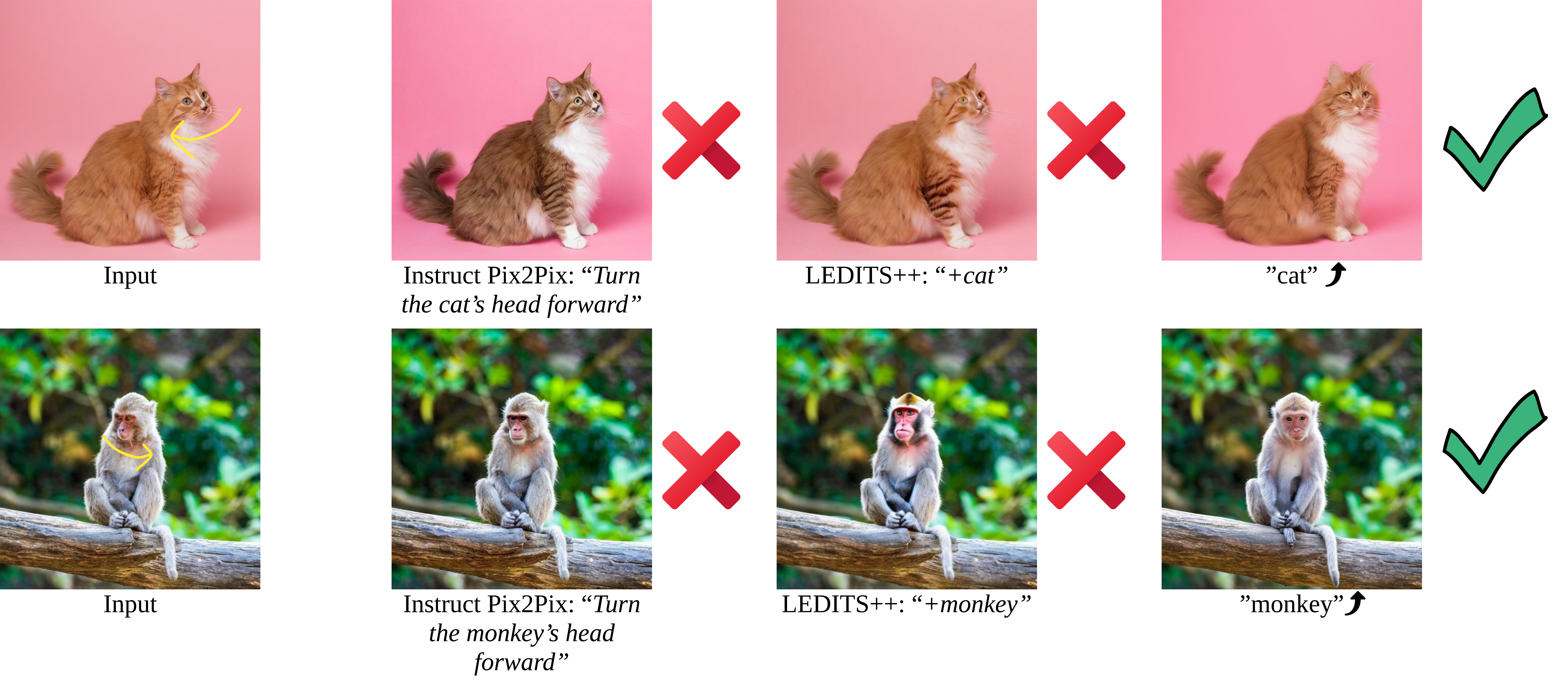}
    \centering
\caption{Given the canonical pose generation effect, we attempt to use Instruction Pix2Pix and LEDITS++ to achieve similar results; however, both approaches failed, demonstrating the challenge of this task.}
\label{fig:pose_ablation}
\end{figure*}

As demonstrated in \cref{fig:pose generation}, our ScalingConcept method can achieve surprising canonical pose generation effects. To further investigate the difficulty of this task, we employ two popular image editing methods: Instruct Pix2Pix~\cite{instructpix2pix}, which follows instructions for editing, and LEDITS++, which adds or removes concepts from the input. Specifically, we instruct Instruct Pix2Pix to ``turn the monkey's head forward,'' but the method fails to produce the desired effect. Similarly, when attempting to add the same concept to the input, LEDITS++ does not achieve the pose generation effect, indicating that this task is non-trivial.

\begin{figure*}[!h]
\centering

    \includegraphics[width=\linewidth]{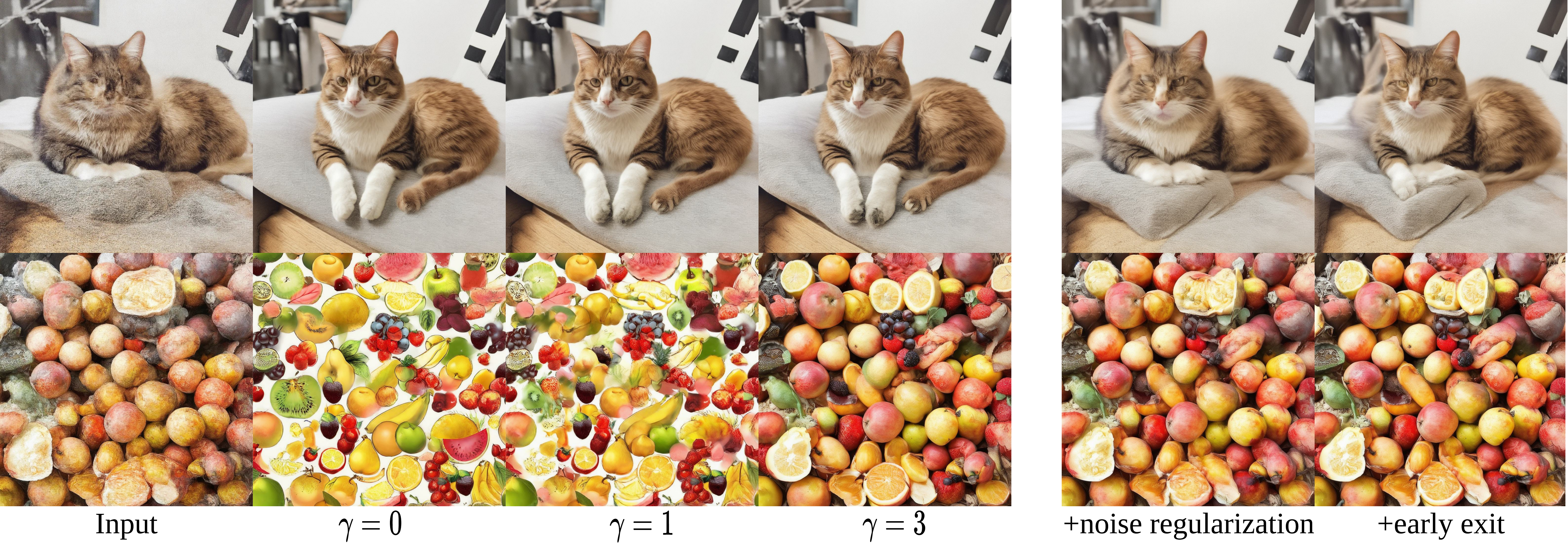}
    \centering
\caption{Visualization of ablation studies. We present the results of concept scaling with different method variants.}
\label{fig:visualization_ablation}
\end{figure*}

\subsection{Visualization of Ablation Studies}
To illustrate the effects of different components of our method, we visualize the results in \cref{fig:visualization_ablation}, which scales up the concepts of ``cat'' and ``fruits'' with $\omega_{base}=5$. The results demonstrate that our non-linear schedule achieves a better trade-off between fidelity and content preservation. Moreover, adding noise regularization helps preserve more fine-grained details, while the introduction of early exit further improves the trade-off.

\begin{figure*}[!h]
\centering

    \includegraphics[width=\linewidth]{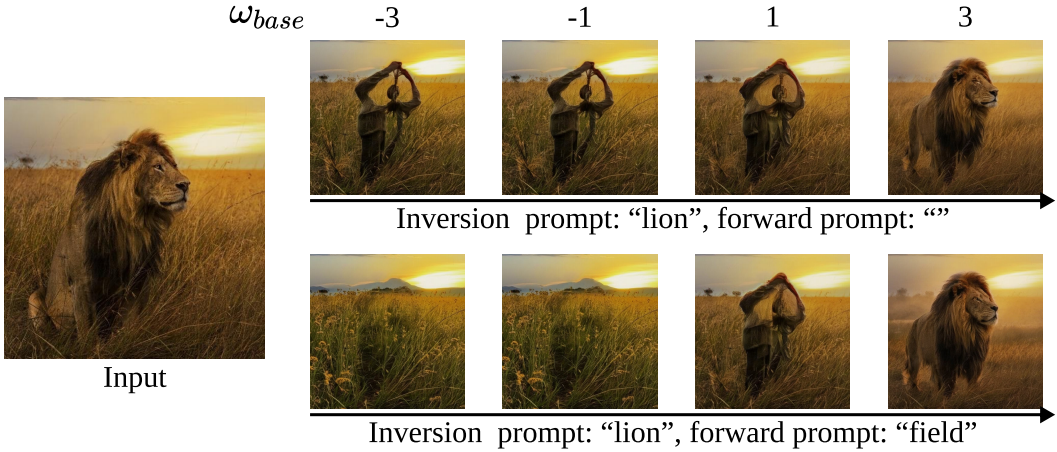}
    \centering
\caption{We set $\gamma=3$ and vary $\omega_{base}$ to investigate its effect. Additionally, we change the prompt from $\emptyset$ to ``field'' to examine the impact of the forward prompt.}
\label{fig:prompt and omega}
\end{figure*}

\subsection{Effect of $\omega_{bsae}$} 
In the previous experiments, we fix $\omega_{base}$ to investigate the effectiveness of other components. In \cref{fig:prompt and omega}, we showcase the effects of varying $\omega_{base}$, with values ranging from -3 to 3, while fixing $\gamma=3$. The figure demonstrates that reducing $\omega_{base}$ corresponds to the removal of the concept, whereas increasing it enhances the concept. However, we found that the removal effect is not as satisfactory as the enhancement, which highlights a limitation related to text-to-image association.

\subsection{Does Forward Prompt Matter?}
In \cref{fig:prompt and omega}, changing the forward prompt from $\emptyset$ to ``field,'' another concept present in the original input, improves the removal effect, as the region left by the null prompt is inpainted with the concept of ``field.'' This demonstrates the importance of selecting the correct concept to serve as the removal helper. However, this approach requires additional effort to label the concepts instead of simply using the versatile null prompt. This suggests an advanced setting for the method, where providing coarse-level annotations for an additional concept can lead to significant improvements.
\end{document}